\documentclass[letterpaper, 10 pt, conference, nofonttune]{ieeeconf}
\IEEEoverridecommandlockouts
\usepackage{microtype}

\usepackage{cite}
\usepackage[bookmarks=true]{hyperref}
\usepackage{color}
\usepackage{times}
\usepackage{epsfig}
\usepackage{epstopdf}
\usepackage{graphicx}
\usepackage{amsmath}
\usepackage{amssymb}
\usepackage{bbm}
\usepackage[ruled,vlined,linesnumbered]{algorithm2e}
\usepackage{graphicx}
\usepackage{verbatim}
\usepackage{booktabs}
\usepackage{multirow}
\usepackage{makecell}
\usepackage{subfigure}
\usepackage{colortbl,booktabs}
\usepackage{lscape}
\usepackage{url}
\usepackage{makecell}
\usepackage{xcolor,colortbl}

\newcommand{\Mult}{\mathbf{\Phi}}
\newcommand{\mult}{\mathbf{\phi}}

\newcommand{\Real}{\mathbb{R}}
\newcommand{\ZZZ}{\mathbf{Z}}
\newcommand{\Zhat}{\hat{\mathbf{Z}}}
\newcommand{\AAA}{\mathbf{A}}

\newcommand{\LLL}{\mathbf{L}}
\newcommand{\One}{\mathbf{1}}
\newcommand{\III}{\mathbf{I}}

\newcommand{\mutwo}{\frac{\mu}{2}}
\newcommand{\onemu}{\frac{1}{\mu}}

\newcommand{\VV}{\mathbf{V}}
\newcommand{\UU}{\mathbf{U}}
\newcommand{\GGG}{\mathcal{G}}
\newcommand{\VVV}{\mathcal{V}}
\newcommand{\EEE}{\mathcal{E}}
\newcommand{\CCC}{\mathcal{C}}

\newcommand{\vneg}{\vspace{-7pt}}

\begin{document}

\title{\LARGE \bf Team Assignment for Heterogeneous Multi-Robot Sensor \\ Coverage through Graph Representation Learning}

\author{
    Brian Reily and Hao Zhang%
    \thanks{*This work was partially supported by NSF CAREER Award IIS-1942056 and NSF CNS-1823245.}%
    \thanks{Brian Reily and Hao Zhang are with the Human-Centered Robotics Lab 
    in the Department of Computer Science
    at the Colorado School of Mines, Golden, CO, 80401.
    Email: \{breily, hzhang\}@mines.edu.}%
}

\maketitle

\begin{abstract}

Sensor coverage is the critical multi-robot problem of maximizing the detection of events in an environment through the deployment of multiple robots. Large multi-robot systems are often composed of simple robots that are typically not equipped with a complete set of sensors, so teams with comprehensive sensing abilities are required to properly cover an area. Robots also exhibit multiple forms of relationships (e.g., communication connections or spatial distribution) that need to be considered when assigning robot teams for sensor coverage. To address this problem, in this paper we introduce a novel formulation of sensor coverage by multi-robot systems with heterogeneous relationships as a graph representation learning problem. We propose a principled approach based on the mathematical framework of regularized optimization to learn a unified representation of the multi-robot system from the graphs describing the heterogeneous relationships and to identify the learned representation's underlying structure in order to assign the robots to teams. To evaluate the proposed approach, we conduct extensive experiments on simulated multi-robot systems and a physical multi-robot system as a case study, demonstrating that our approach is able to effectively assign teams for heterogeneous multi-robot sensor coverage.

\end{abstract}

\section{Introduction}
\label{sec:intro}

Multi-robot sensor coverage is a critical problem for multi-robot systems,
with the objective of deploying a multi-robot team 
in an area in order to maximize the overall sensing
performance in terms of the detection of
phenomena or events in the environment 
\cite{cortes2004coverage,luo2018adaptive,schwager2006distributed}.
Multi-robot sensor coverage enables collaborative observations
of large and complex environments,
which allows a multi-robot system to effectively obtain a 
more complete view of the environment than each 
individual robot could.
Multi-robot sensor coverage is the core task in a wide 
variety of real-world applications,
including
surveillance \cite{meguerdichian2001exposure}, 
search and rescue \cite{zadorozhny2013information},
and environment exploration \cite{luo2016distributed}.

Multi-robot sensor coverage is a challenging research problem.
In real-world environments, e.g., during disaster response,
various phenomena or events can occur.
However, 
as individual robots in a large multi-robot system are typically limited in
their sensing, mobility, and computation capabilities \cite{parker1994heterogeneous},
individual robots are often not be able to cover
an entire area or may not have the sensing capabilities to sense all events in the environment.
Therefore, these robots must be organized into teams in a such way that 
the maximum number of phenomena are detected through distributing their
available sensing capabilities \cite{santos2018coverage}.
Furthermore, robots in a multi-robot system exhibit 
multiple additional heterogeneous relationships, such as
their communication connections and spatial 
distribution \cite{reily2020representing},
which must be taken into account when assigning
teams for sensor coverage in a multi-robot system.

Due to its importance, multi-robot sensor coverage has been
attracting significant attention over the past few years. 
Many previous approaches have focused on sensor coverage by homogeneous multi-robot
systems \cite{cortes2004coverage,pierson2017adapting}.
These techniques assumed that the robots possess the same set of sensors
and cannot address heterogeneity in sensing capabilities. 
To address heterogeneous sensing abilities, 
several methods have been implemented to assign coverage
regions based on Gaussian distributions \cite{luo2018adaptive},
control laws \cite{santos2018coverage},
and information maximization \cite{sadeghi2019coverage}.
The methods consider only the spatial distribution of robots or only
the sensing capabilities, 
and cannot integrate multiple heterogeneous relationships
that occur in a multi-robot system when performing sensor coverage.

In this paper, we propose an approach to assigning teams for
sensor coverage by 
a multi-robot system with multiple heterogeneous relationships 
through novel graph representation learning.
We describe each of the heterogeneous relationships among the robots
(such as their spatial distribution, communication connections,
and co-occurrence of sensing capabilities)
as a graph that is encoded using an adjacency matrix.
Then, we formulate team assignment as a 
graph representation learning problem,
and develop a method to learn a unified representation 
that integrates multiple graphs describing the heterogeneous relationships.
The proposed approach is based on the principled mathematical
framework of regularized optimization
with structured norms as regularization terms in order to identify block structures within the representation.
The learned representation is used to perform sensor coverage
by assigning the robots to teams according to a given number of regions based on spectral cuts.

This paper introduces two contributions:
\begin{itemize}
\item We introduce a new problem formulation that 
formulates sensor coverage by a multi-robot system having multiple heterogeneous relationships
as a problem of assigning teams through graph representation learning.
We also propose a novel principled approach based
upon regularized optimization
which learns a unified representation of the 
multi-robot system 
from the graphs describing the heterogeneous robot relationships,
and applies structured norms and constraints to identify the 
underlying structure 
within the representation.
\item We develop an iterative algorithm to solve the formulated
regularized optimization problem, 
which is challenging to solve because of the non-smooth regularization terms and constraints.
We prove that this algorithm is theoretically 
guaranteed to converge to the optimal solution.
\end{itemize}

\section{Related Work}
\label{sec:related}


\subsection{Homogeneous Multi-Robot Sensor Coverage}

Most research has focused on multi-robot systems with
homogeneous sensing capabilities \cite{cortes2004coverage}.
As approaches designed for homogeneous multi-robot systems
assume only the type of sensing modality, 
most methods address
sensor coverage from the perspective of fully covering an
environment spatially, and devising multi-robot strategies to do this
efficiently \cite{rekleitis2008efficient}.
Partitioning a space based on the estimated information gain from
different regions was proposed in \cite{fung2019coordinating},
while \cite{corah2017efficient} introduced a distributed version
of sequential greedy assignment to plan coverage paths.
Representing an environment as a discrete graph and identifying
equal-mass partitions to assign to each individual robot was
used in \cite{yun2014distributed,yun2012distributed}.
Deployment of multiple robots to cover an unknown environment
is approached
as a problem of distributing robots in the
environment \cite{cortes2002coverage}.
This can be done by using gradient descent over estimated
density functions \cite{lee2015multirobot} or assigning Voronoi
partitions \cite{guruprasad2010automated,guruprasad2013performance}.


All of these described methods apply to only homogeneous teams, where
each robot is assumed to have the same sensing capabilities.
Thus, they are not able to address sensor coverage by heterogeneous robots 
with various capabilities.

\subsection{Heterogeneous Multi-Robot Sensor Coverage}

For small multi-robot teams,
approaches to sensor coverage based on
scheduling \cite{manjanna2018heterogeneous,shkurti2012multi} or
naive following (e.g., an aerial robot follows a ground robot to
provide a different perspective) \cite{hood2017bird} have been
effective.
For larger multi-robot systems, without strictly defined roles or
capabilities, more general approaches are necessary.
Some again use Voronoi partitions, 
assigning each robot a specific region to
cover \cite{arslan2016voronoi,guruprasad2013heterogeneous}.
Other methods have been focused on 
identifying environment correspondences based on robot
locations \cite{gao2020correspondence,gao2020regularized} or
fitting robot positions to a
distribution function based on `sensing quality' \cite{sadeghi2019coverage}.
Identifying this distribution from the sensors in the environment
and dynamically responding to it has also been proposed,
through identifying the most informative areas \cite{schwager2006distributed}
or by fitting a density function to an exact sensed value
such as temperature \cite{luo2018adaptive}.
This has also been addressed using weighted density functions that are
adjusted as the robots sense more of the environment and estimate
the true density \cite{schwager2009decentralized} or
by utilizing gradient descent based methods to converge to
locally optimal arrangements \cite{le2012adaptive}.
The idea of evaluating multi-robot sensor coverage based on the detection
of multiple event types was introduced to rate a cost function
to distribute robots \cite{santos2018coverage,santos2018coverage_b}.

While these approaches have been implemented to assign regions
or tasks to individual robots in a multi-robot system (e.g.,
through game theory \cite{jang2017anonymous,jang2018comparative} or
scheduling algorithms \cite{brutschy2014self,castillo2014temporal}),
little existing research has focused on the problem of
identifying teams of robots which would work together to perform sensor coverage.
Naive methods have been introduced that rely solely on
line-of-sight \cite{rekleitis2004limited} or identifying
equal sized working regions \cite{schneider1998territorial}.
Other existing methods are biologically inspired by real-world
insect behavior, such as ant foraging \cite{labella2006division},
ant colony labor division \cite{wu2018modeling},
or swarms of wasps \cite{bonabeau1997adaptive}.
However, these methods are limited by either being overly
parametric (in order to imitate a complex existing biological model)
or simplistic (e.g., by assuming the agents would not be
able to communicate).

Although these described approaches are able to deal with multiple
sensing capabilities,
they are based upon single forms of relationships 
(typically
spatial relations among robots in the environment)
and 
cannot integrate multiple heterogeneous relationships,
while real-world multi-robot systems are typically defined through
multiple forms of heterogeneous relationships.


\section{Our Proposed Formulation and Approach}
\label{sec:approach}




\subsection{Problem Formulation}

Given a multi-robot system at a distinct time point 
consisting of $N$ robots with $M$ heterogeneous relationships,
we can describe each relationship
using a directed graph $\GGG_m = \left( \VVV, \EEE_m \right)$,
where $\VVV = \{ v_1, \dots, v_N \}$ denotes the set of vertices and
$\EEE_m$ denotes the set of directed edges between these vertices for the
$m$-th relationship.
When describing the multi-robot system, each vertex $v_i$ represents an individual robot.
Each edge $e_{ij} = \left( v_i, v_j \right) \in \EEE_m$ represents the connection
between the robots corresponding to vertices $v_i$ and $v_j$ in the
$m$-th relationship.
The magnitude and direction of the edge $e_{ij}$ depends on the relationship that
is being modeled, e.g. a spatial relationship may be represented
by the distance between the two robots.
Each graph $\GGG_m$ is then encoded using a corresponding adjacency matrix $\AAA_m \in \Real^{N \times N}$,
with each element $a_{ij}$ describing the value of the edge $e_{ij}$.
For example, if a graph $\GGG_m$ encodes the co-occurrence of sensing capabilities
in a multi-robot system, the edge $e_{ij}$ and the 
entry $a_{ij}$ in the adjacency matrix could have a real value of the number of sensors in common
between the $i$-th robot and the $j$-th robot.
Assuming these robots have $M$ heterogeneous relationships,
the multi-robot system can be represented by the $M$-order graph
$\GGG = \left( \VVV, \EEE_1, \dots, \EEE_M \right)$.

We formulate team assignment for heterogeneous multi-robot systems
as a graph representation learning problem,
with the objective of learning a unified representation of 
the multi-robot system from the $M$-order graph $\GGG$,
which can be used to assign robots into teams.
This formulation can be formally defined as an optimization problem.
Given $\GGG$,
the goal is to obtain a
matrix $\ZZZ = \{ z_{ij} \} \in \Real^{N \times N}$ that optimally 
aggregates the individual graphs in $\GGG$, where each element
$z_{ij}$ describes the probability that the $i$-th robot and the
$j$-th robot should be assigned to the same team and coverage region.
This matrix $\ZZZ$ represents an adjacency matrix that
approximates the overall structure of the multi-robot system.
Mathematically, we learn $\ZZZ$ via a loss function,
$\min_{\ZZZ} \mathcal{L} \left( \AAA ; \ZZZ \right)$,
where the loss is based on 
how well $\ZZZ$ approximates
each individual graph in $\GGG$:
\begin{align}
\min_{\ZZZ} & \sum_{m=1}^{M} \alpha_m \| \ZZZ - \AAA_m \|_F^2
\end{align}
where $\|\cdot\|_F$ denotes the Frobenius norm, 
and $\alpha_m$, for $m = 1, \dots, M$ are hyperparameters
where $\sum_{m=1}^{M} \alpha_m = 1$, 
which are used to control the influence of each graph
and can be tuned via grid search or be defined by experts.

\subsection{Identifying Structures of Multi-Robot Systems}

Based on this problem formulation, 
we identify underlying block structures within 
the unified representation matrix $\ZZZ$ that
correspond to teams for sensor coverage.
We enforce $\ZZZ$ to be bistochastic (i.e.,
a non-negative real matrix whose rows and columns all sum to 1).
We enforce this for three reasons and introduce three corresponding
constraints to the structure of $\ZZZ$.
First, as each element $z_{ij}$ of $\ZZZ$ describes a probability relationship
between robot $i$ and robot $j$ (i.e., the probability that the robots
should be teamed together), no element of $\ZZZ$ can be less than $0$,
since a probability must be positive.
For this, we add the constraint that $\ZZZ \geq 0$.
Second, these probabilistic relationships are reflexive, as the probability
that the $i$-th robot should be teamed with the $j$-th robot should
be the same as the probability that the $j$-th robot should be teamed
with the $i$-th robot.
Because of this, $z_{ij}$ should equal $z_{ji}$, so we
add the constraint that $\ZZZ = \ZZZ^T$.
Third, because each element $z_{ij}$ describes a
probabilistic relationship between the $i$-th robot
and the $j$-th robot, each row $\mathbf{z}^i$ and
each column $\mathbf{z}_j$ should sum to $1$, as the total connection
probability for each individual robot should also sum to $1$.
Because of this, we add the constraint that $\ZZZ \One = \One$, where
$\One$ denotes a vector of 1s.
With these constraints, our formulation becomes:
\begin{align}
\label{eq:bistoch}
\min_{\ZZZ} & \sum_{m=1}^{M} \alpha_m \| \ZZZ - \AAA_m \|_F^2 \\
\text{s.t.} & \; \ZZZ \One = \One, \ZZZ = \ZZZ^\top, \ZZZ \geq 0 \notag
\end{align}

We now identify teams within the multi-robot system 
by inducing a learned block structure
within $\ZZZ$ through designing new structured sparsity-based regularization
terms,
which can be integrated into Eq. (\ref{eq:bistoch}) in order
to regularize $\ZZZ$ under the mathematical framework of regularized optimization.

The first regularization term we introduce utilizes the squared Frobenius norm
on the $\ZZZ$ matrix:
\begin{align}
\| \ZZZ \|_F^2 = {\sum_{i=1}^{N} \sum_{j=1}^{N} z_{ij}^2}
\end{align}
This norm penalizes high values in $\ZZZ$, restricting
connection probabilities to small numbers of vertices.
Through enforcing our bistochastic constraint that all rows and columns
must sum to 1, yet penalizing high values with this norm, we increase the
probabilities among strongly connected robots, while
causing probabilities between weakly connected robots to approach 0.

The second regularization term we introduce acts on the spectrum of the
unified representation matrix $\ZZZ$,
in the form of the nuclear norm on the Laplacian of
the $\ZZZ$ matrix.
Because of the bistochastic constraints introduced in
Eq. (\ref{eq:bistoch}), we know that each
row and column of $\ZZZ$ sum to 1, and thus the degree matrix of $\ZZZ$,
representing the total in-degree and out-degree of each vertex, is equal
to the identity matrix $\III$.
Thus we can define the Laplacian of $\ZZZ$ as $\LLL = \III - \ZZZ$.
The nuclear norm of a matrix is equivalent to the $\ell_1$-norm of
that matrix's singular values, or the square roots of the matrix's
eigenvalues.
Thus it encourages sparsity among these eigenvalues,
penalizing the terms which are non-zero.
As the multiplicity of $0$ eigenvalues of a Laplacian matrix corresponds
to a graph's connectivity (i.e., the Laplacian of a fully connected
graph has a single $0$ eigenvalue),
enforcing sparsity
by way of the nuclear norm encourages the formation of a
graph with more connected components:
\begin{align}
\| \LLL \|_* = \sum_{n=1}^{N} \sigma_n \left( \LLL \right)
\end{align}

\begin{algorithm}[tb]
\small
\SetAlgoLined
\SetKwInOut{Input}{Input}
\SetKwInOut{Output}{Output}
\SetNlSty{textrm}{}{:}
\SetKwComment{tcc}{/*}{*/}
\BlankLine
\vspace{-3pt}

Set $1 < \rho < 2$ and $k = 0$. 
Initialize the penalty terms $\mu^0$, $\mult_1^0$, $\Mult_2^0$,
$\Mult_3^0$, and $\Mult_4^0$.

\Repeat{
convergence
}{
Compute $\ZZZ^{k+1}$ by Eq. (\ref{eq:update_z}).

Compute $\Zhat^{k+1}$ by Eq. (\ref{eq:update_zhat}).

Compute $\LLL^{k+1}$ by Eq. (\ref{eq:update_l}).

Update $\mult_1$ by $\mult_1^{k+1} = \mult_1^k + \mu^k \left( \ZZZ^{k+1} \One - \One \right)$.

Update $\Mult_2$ by $\Mult_2^{k+1} = \Mult_2^k + \mu^k \left( {\ZZZ^{k+1}}^\top - \Zhat^{k+1} \right)$.

Update $\Mult_3$ by $\Mult_3^{k+1} = \Mult_3^k + \mu^k \left( \LLL^{k+1} - \III + \ZZZ^{k+1} \right)$.

Update $\Mult_4$ by $\Mult_4^{k+1} = \Mult_4^k + \mu^k \left( \Zhat^{k+1} - \ZZZ^{k+1} \right)$.

Update $\mu$ by $\mu^{k+1} = \rho \mu^k$.

$k = k + 1$.
}
\vspace{-1pt}
\caption{Our Algorithm to Solve Eq. (\ref{eq:opt_final}).}
\label{alg:alm2}
\end{algorithm}

With the Frobenius norm concentrating values among small groups of
vertices and the nuclear norm of the Laplacian rewarding graphs
with more connected components,
we cause blocks to form in $\ZZZ$, corresponding to
the underlying structure of the multi-robot system.
Using both norms as regularization terms in the objective function allows this underlying structure
to emerge by focusing the loss function on accurately
representing the individual graphs but rewarding the formation of groups
within the unified representation.
When $\GGG$  has an underlying structure that contains $k$ groupings, with rearrangement
of rows and columns $\ZZZ$ will appear as a block matrix with $k$ blocks.

With the two regularization terms, our final formulation is defined as 
a regularized constrained optimization problem:
\begin{align}
\label{eq:opt_final}
\min_{\ZZZ} & \sum_{m=1}^{M} \alpha_m \| \ZZZ - \AAA_m \|_F^2  + \lambda_1 \| \ZZZ \|_F^2 + \lambda_2 \| \LLL \|_* \\
\text{s.t.} & \; \LLL = \III - \ZZZ, \ZZZ \One = \One, \ZZZ = \ZZZ^\top, \ZZZ \geq 0 \notag
\end{align}
As the nuclear norm is not convex and $\LLL$ and $\ZZZ$ are dependent,
we design a new iterative algorithm, as presented by Algorithm \ref{alg:alm2}, 
to solve the formulated optimization problem and obtain the optimal $\ZZZ$.
The algorithm will be detailed in Section \ref{sec:opt}.

\subsection{Assigning Multi-Robot Teams for Sensor Coverage}

We address the problem of multi-robot sensor coverage by assigning
robots to coverage regions based on the learned relationships
represented in $\ZZZ$.
As $\ZZZ$ is learned from the heterogeneous
relationships that describe the multi-robot system, the blocks
induced by the introduced structured sparsity-inducing regularization terms
correspond to teams of robots that are
located near each other while together possessing a variety of sensing
capabilities.

We continue to treat $\ZZZ$ as an adjacency matrix corresponding
to the learned unified representation of the multi-robot system.
The algebraic connectivity of a graph is defined as the second
smallest eigenvalue of the Laplacian, also referred to as the Fiedler
value.
The corresponding eigenvector, known as the Fiedler vector, can
be utilized to partition a graph based upon the signs of its values
\cite{fiedler1973algebraic}.
We begin the process of applying cuts based on the Fiedler vector to
cover a set of regions by defining $r$, the mission-dependent number of
regions to be covered.
We then iteratively apply Fiedler cuts to $\ZZZ$ until we have $r$
partitions.
We do this by first partitioning based upon the entire $\ZZZ$ matrix,
and then subsequently partitioning based on minors of $\ZZZ$,
by re-partitioning the largest existing grouping.
For example, if $\ZZZ$ is $\in \Real^{5 \times 5}$ and initially
partitions into $\{ 1, 2 \}$ and $\{ 3, 4, 5 \}$, the next partition
is done on the minor of $\ZZZ$ based on the overlap of the
$\{ 3, 4, 5 \}$ columns and the $\{ 3, 4, 5 \}$ rows.
These partitions correspond to the assignment of
multi-robot teams, e.g. in this case robots $1$ and $2$ would
be teamed together if no further cuts were made.

\section{Optimization Algorithm}
\label{sec:opt}

The constrained optimization problem in our final formulation in Eq. (\ref{eq:opt_final}) is hard
to solve, mainly because the nuclear norm is not convex and
because of the dependency between $\LLL$ and $\ZZZ$.
To solve it, we introduce a solution based on the 
Augmented Lagrange Multiplier (ALM) method, which solves
problems of the form
\begin{align}
\text{min} \; & f \left( \mathbf{X} \right) \; \text{s.t.} \; h \left( \mathbf{X} \right) = 0 \notag
\end{align}
by rewriting constraints as penalty terms.
We introduce $\Zhat = \ZZZ$ and solve for
$\ZZZ$, $\Zhat$, and $\LLL$ iteratively 
to converge to a final solution.
We also introduce $\mu, \mult_1, \Mult_2, \Mult_3$, and $\Mult_4$ 
and are able to rewrite our equality constraints as
penalty terms:
\begin{align}
\label{eq:opt_with_penalties}
\min_{\ZZZ} & \sum_{m=1}^{M} \alpha_m \| \ZZZ - \AAA_m \|_F^2  + \lambda_1 \| \ZZZ \|_F^2 + \lambda_2 \| \LLL \|_* \\
& + \mutwo \| \ZZZ \One - \One + \onemu \mult_1 \|_2^2 + \mutwo \| \ZZZ^\top - \Zhat + \onemu \Mult_2 \|_F^2 \notag \\
& + \mutwo \| \LLL - \III + \ZZZ + \onemu \Mult_3 \|_F^2 + \mutwo \| \Zhat - \ZZZ + \onemu \Mult_4 \|_F^2 \notag \\
\text{s.t.} & \; \ZZZ \geq 0 \notag
\end{align}

Our iterative algorithm is defined in Algorithm \ref{alg:alm2}
and described below.
\subsubsection{\textbf{Step 1}}
We first solve for $\ZZZ$, by fixing $\Zhat$ and $\LLL$ and taking
the derivative of Eq. (\ref{eq:opt_with_penalties}) w.r.t. $\ZZZ$
and setting it equal to 0.
The update to $\ZZZ$ at each iteration is
\begin{align}
\ZZZ = & \; \Bigg( \sum_{m=1}^M 2 \alpha_m \AAA_m + \mu \left( \One \One^\top + \Zhat^\top + \LLL - \III - \Zhat \right) \notag \\
& \; \; \; \; - \mult_1 \One^\top - \Mult_2 + \Mult_3 + \Mult_4 \Bigg) \notag \\
& \Bigg( \sum_{m=1}^M 2 \alpha_m \III + 2 \lambda_1 \III + 3 \mu \III + \mu \One \One^\top \Bigg)^{-1}
\end{align}
To incorporate the $\ZZZ \geq 0$ constraint, the final update to
$\ZZZ$ is
\begin{align}
\label{eq:update_z}
\ZZZ = \text{max} \{ \ZZZ, 0 \}
\end{align}

\subsubsection{\textbf{Step 2}}
Next, we solve for $\Zhat$, by again fixing the other two variables
and taking the derivative of Eq. (\ref{eq:opt_with_penalties})
w.r.t. $\Zhat$ and setting it equal to 0.
With rearrangement, the final update to $\Zhat$ is
\begin{align}
\label{eq:update_zhat}
\Zhat = \left( 2 \mu \III \right)^{-1} \left( \mu \ZZZ^\top + \mu \ZZZ + \Mult_2 + \Mult_4 \right)
\end{align}

\subsubsection{\textbf{Step 3}}
Next, we solve for $\LLL$ by minimizing the partial objective
function
\begin{align}
\min_\LLL & \; \lambda_2 \| \LLL \|_* + \mutwo \| \LLL - \III + \ZZZ + \onemu \Mult_3 \|_F^2
\end{align}

We solve this by computing the singular value decomposition (SVD) of 
$-\III + \ZZZ + \frac{\Mult_3}{\mu}$:
\begin{align}
\text{SVD} \left( -\III + \ZZZ + \frac{\Mult_3}{\mu} \right) = \UU \Sigma \VV^\top
\end{align}
and update $\LLL$ at each iteration by
\begin{align}
\label{eq:update_l}
\LLL = \UU \text{diag} \left( \left( \sigma_i - \frac{\lambda_2}{\mu} \right)_{+} \right) \VV^\top
\end{align}
where $\sigma_i$ is the $i$-th diagonal element of $\Sigma$ and
$\text{diag} \left( \mathbf{x} \right)$ is a diagonal matrix with the
elements of $\mathbf{x}$ on the diagonal.

\subsubsection{\textbf{Step 4}}
Finally, we update the $\mu$, $\mult_1$, $\Mult_2$, $\Mult_3$, $\Mult_4$
and $k$ parameters by the equations in Lines 6 through 10.

We obtain the optimal solution by repeating the update 
to $\ZZZ$ in Line 3, $\Zhat$ in Line 4, $\LLL$ in Line 5,
and the updates to the multiplier variables in Lines 6--10
until convergence.


\begin{figure*}[!t]
    \centering
    \subfigure[20 Robots, 3 Regions]{
        \label{fig:vor_a}
        \centering
        \includegraphics[width=0.22\textwidth]{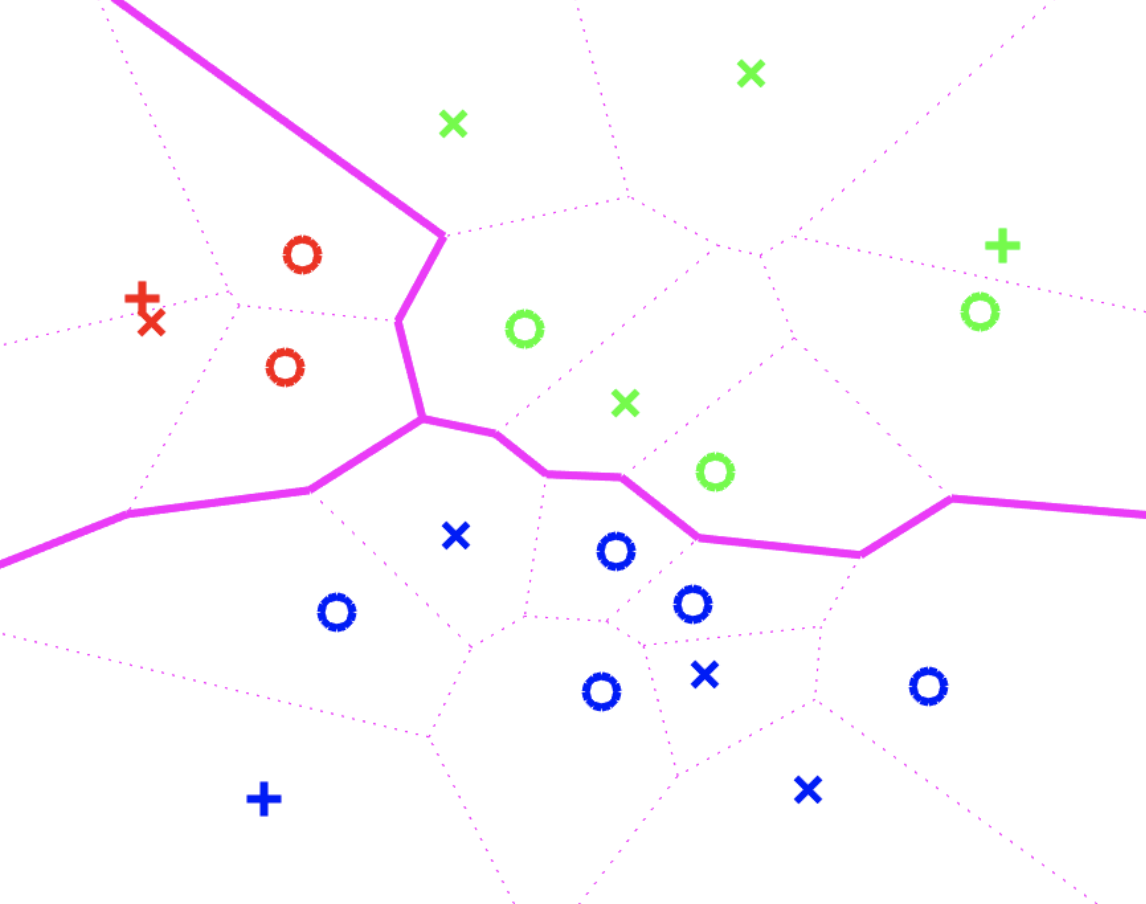}
    }%
    \subfigure[40 Robots, 3 Regions]{
        \label{fig:vor_b}
        \centering
        \includegraphics[width=0.22\textwidth]{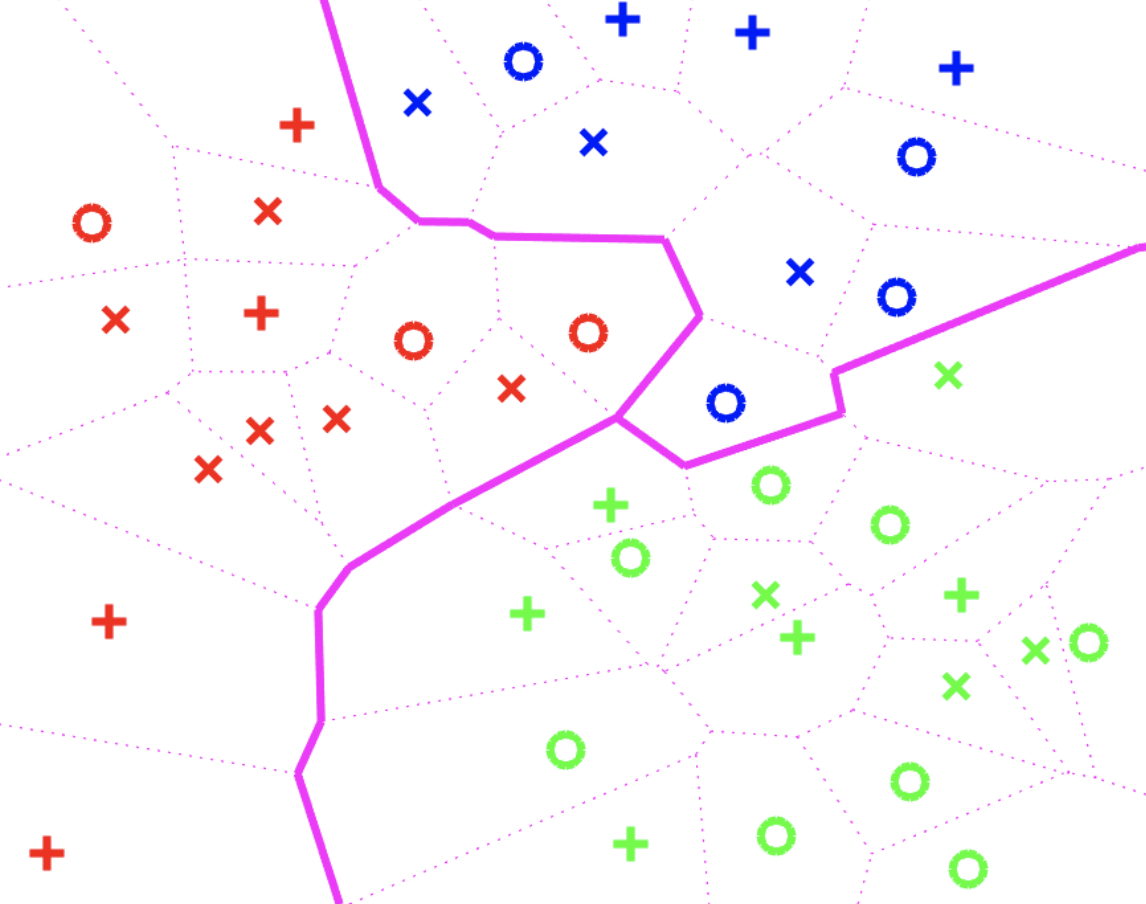}
    }%
    \subfigure[20 Robots, 3 Regions]{
        \label{fig:obs_b}
        \centering
        \includegraphics[width=0.22\textwidth]{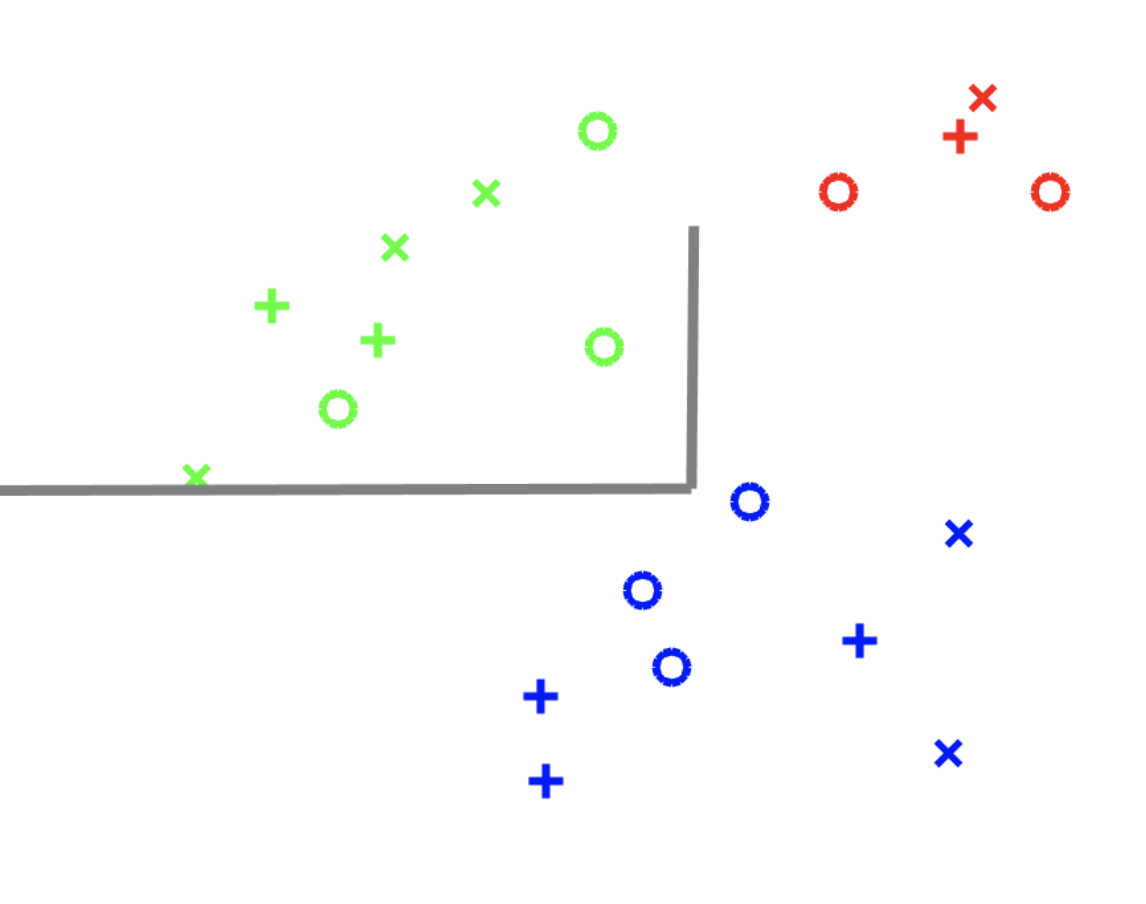}
    }%
    \subfigure[20 Robots, 4 Regions]{
        \label{fig:obs_c}
        \centering
        \includegraphics[width=0.22\textwidth]{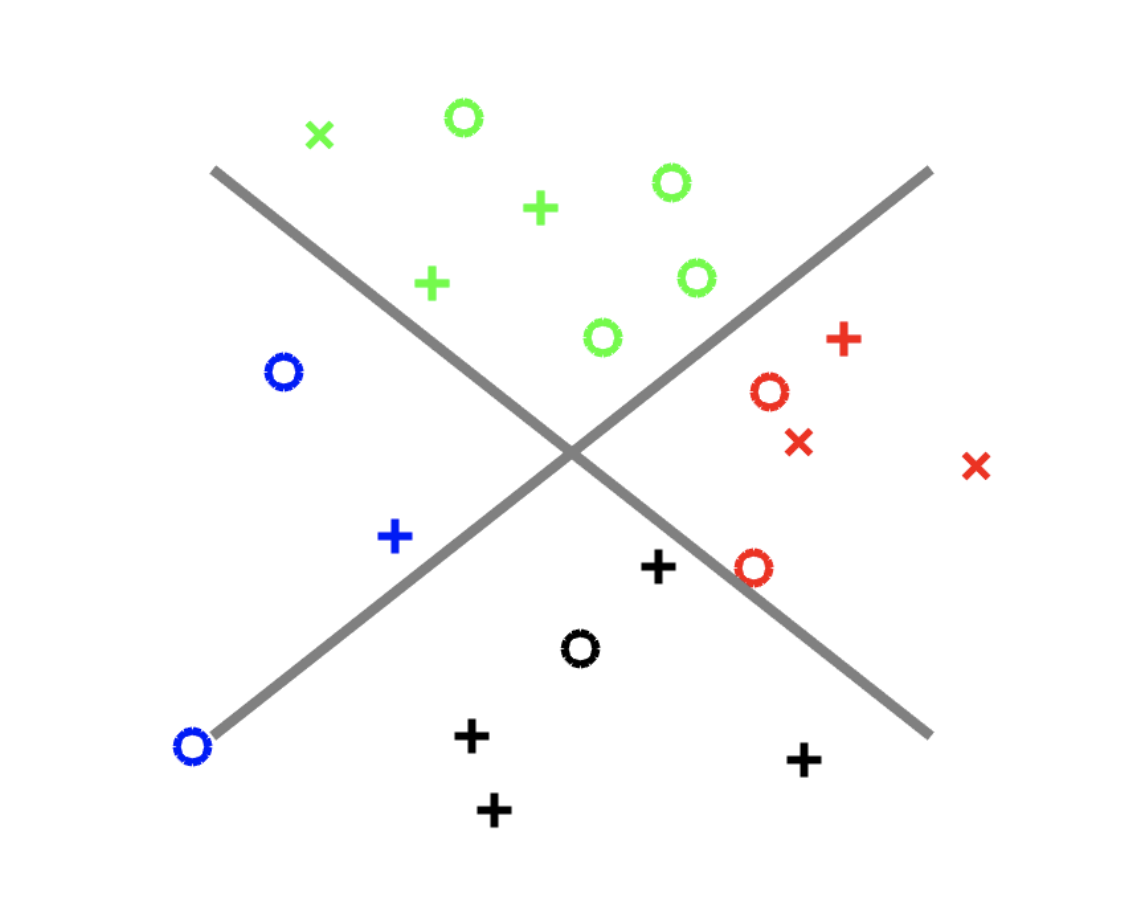}
    }
    \vneg
    \caption{Figures \ref{fig:vor_a} and \ref{fig:vor_b} show
    coverage regions for varying number of robots.
    Light dotted magenta lines show the spacing of individual robots,
    while bold magenta lines depict the coverage regions of
    the heterogeneous teams.
    Figures \ref{fig:obs_b} and \ref{fig:obs_c} show 
    our approach's performance with obstacles in the environment
    (marked with gray lines).
    Voronoi regions are not marked here to reduce visual noise.
    Markers show individual robots, 
    with the marker color
    denoting the team assignment and the marker shape indicating
    that robot's capability (e.g., a robot marked by an `o' has
    different sensing capability than a robot marked by an
    `x').}
    \label{fig:vor}
    \vneg
\end{figure*}

\subsubsection*{\bf Convergence and complexity}
The general ALM method described and on which our solution is based
is proven to converge to an optimal
solution \cite{bertsekas2014constrained} as long as
$0 < \mu^k < \mu^{k+1}$ for iteration $k$.
With this assumption, the current solution $\mathbf{X}^k$ will approach
the optimal solution $\mathbf{X}^{k*}$.
Since we initialize $\mu^0$ to be positive, initialize $\rho$ such
that $1 < \rho < 2$, and update $\mu$ by $\mu^{k+1} = \rho \mu^k$
(Line 10 in Algorithm \ref{alg:alm2}),
this assumption will hold at every iteration.
In terms of complexity, 
it is trivial to update
$\mult_1$, $\Mult_2$, $\Mult_3$, and $\Mult_4$ in Lines 6-9,
as well as $\mu$ and $k$ in Lines 10 and 11.
The time complexity of our solution is dominated by the update
of $\ZZZ$ in Line 3, $\Zhat$ in Line 4, and $\LLL$ in Line 5.
Lines 3 and 4 compute a matrix inverse and 
multiplication, each of complexity $\mathcal{O} ( N^3 )$.
Line 5 calculates the SVD of a square matrix, also with a complexity of
$\mathcal{O} \left( N^3 \right)$.
As a result, the overall complexity of our solution
is $\mathcal{O} \left( N^3 \right)$.

\section{Experimental Results}
\label{sec:results}

We evaluated on both simulated and physical multi-robot
systems, each described with three different graphs
representing their heterogeneous relationships.
\begin{enumerate}
    \item \emph{Spatial relationship}: The first graph, $\GGG_S$, describes the spatial
    structure of the system.
    Each edge $e_{ij}$ describes the inverse of the distance between the
    $i$-th and the $j$-th robot, causing nearby robots to have 
    higher edge weights than further apart robots.
    \item \emph{Communication}: The second graph, $\GGG_{C}$, describes the
    communication capabilities of the system.
    Robots are physically 
    limited in their communication capabilities, 
    just as they are limited in their sensing capabilities.
    For this graph, $e_{ij} = 1$ if the $i$-th robot is able to
    communicate to the $j$-th robot, and $0$ if not.
    \item \emph{Heterogeneous sensing capability}: The third graph, $\GGG_{SC}$, describes
    the relationships between robots based on their sensing capabilities.
    We define a set of sensing capabilities $\CCC$, with $\CCC_i$ denoting the capabilities of the $i$-th robot.
    $\GGG_{SC}$ represents the similarity of two robots based on their
    relative sensing capabilities.
    Here, $e_{ij} = | \CCC_i \cap \CCC_j |$, or the size of the
    intersection between the $i$-th robot's capabilities and the
    $j$-th robot's capabilities.
\end{enumerate}

\begin{figure*}[!t]
    \centering
    \subfigure[20 Robots, 3 Capabilities]{
        \label{fig:ed_sd_A}
        \centering
        \includegraphics[width=0.22\textwidth]{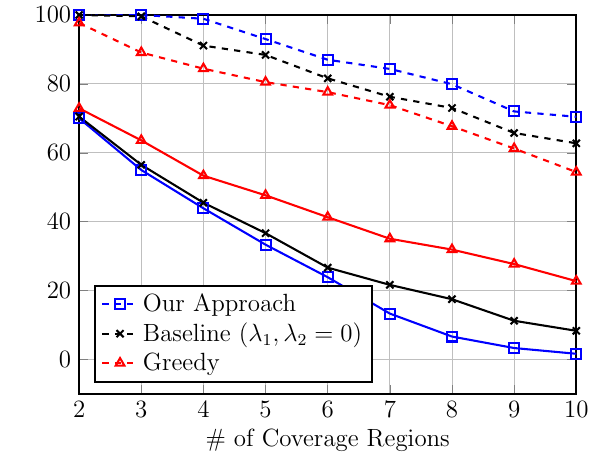}
    }%
    \subfigure[20 Robots, 5 Capabilities]{
        \label{fig:ed_sd_B}
        \centering
        \includegraphics[width=0.22\textwidth]{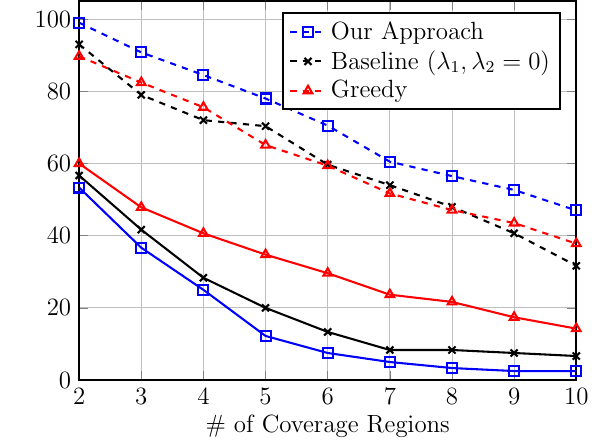}
    }
    \subfigure[40 Robots, 3 Capabilities]{
        \label{fig:ed_sd_C}
        \centering
        \includegraphics[width=0.22\textwidth]{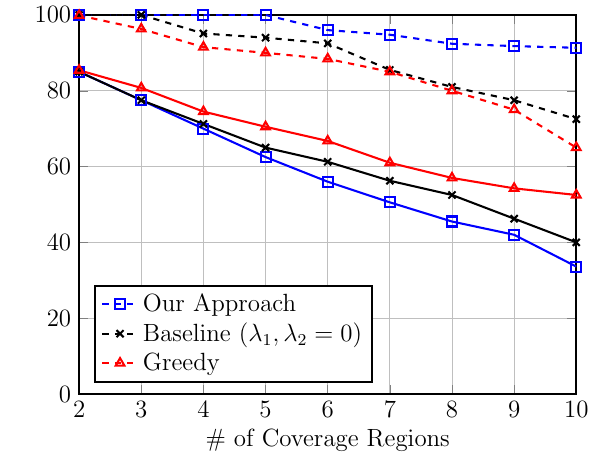}
    }%
    \subfigure[40 Robots, 5 Capabilities]{
        \label{fig:ed_sd_D}
        \centering
        \includegraphics[width=0.22\textwidth]{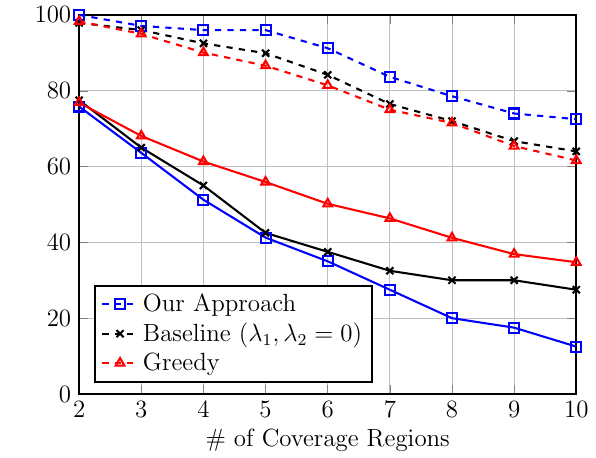}
    }
    \vneg
    \caption{Event detection and robot duplication as the number of coverage regions
    varies, for simulated multi-robot systems of $N = \{ 20, 40 \}$ and
    $| \mathcal{C} | = \{ 3, 5 \}$.
    Dashed lines indicate event detection, where higher rates
    are the better result, while solid lines indicate robot
    duplication, where lower rates are better.}
    \label{fig:ed_sd}
    \vneg
\end{figure*}

We adopt two metrics commonly used in multi-robot sensor coverage literature for evaluation, as well as a new metric to quantitatively evaluate
the composition of our identified teams.
\begin{enumerate}
    \item \emph{Voronoi diagrams} are adopted as a metric to evaluate the coverage regions assigned to each
    heterogeneous team.
    Voronoi regions divide an area into polygons based on the distance
    to different points, and are commonly used to qualitatively assess
    robot coverage methods \cite{luo2018adaptive,santos2018coverage}.
    \item \emph{Event detection} is used to evaluate sensing quality.
    For each testing iteration we simulate 100 events of multiple types,
    each of which is detectable by a single sensing capability,
    following \cite{sadeghi2019coverage,santos2018coverage}.
    If an event occurs in a region
    where a robot team member has the sensing ability to sense that type of event,
    then the event is considered detected.
    If no such robot exists, then the event is
    not detected.
    An event detection rate of $100\%$ is optimal, meaning all events are detected.
    \item \emph{Robot duplication} is introduced
    to evaluate the composition of
    the identified teams.
    If a team contains two robots that share a sensor
    capability, then one robot is a duplicate.
    The sensor duplication score is $\frac{d}{n}$, where $d$ is
    the number of duplicate robots and $n$ is the total number
    of robots.
    Lower duplication rates are better, as this corresponds
    to robots being more evenly distributed.
\end{enumerate}

We compare our approach to a baseline version as well as a
greedy algorithm.
The baseline version of our approach sets $\lambda_1 = \lambda_2 = 0$,
so that our learned representation matrix $\ZZZ$ is still
constrained to be bistochastic but does not utilize the two
regularization terms that induce a block structure.
The greedy algorithm disregards sensor capability
relationships and assigns robots to a region purely based on
spatial relationships, i.e. robots near each other are 
assigned to the same team.

\subsection{Results on Simulated Multi-Robot Systems}

We simulated systems of
$N = \{ 10, 20, 30, 40, 50 \}$
robots, and randomly assign a varying number of sensing capabilities
for $| \mathcal{C} | = \{ 2, 3, 4, 5 \}$.
The graphs describing each system were defined
based on the generated positions and capabilities.

Figures \ref{fig:vor_a} and \ref{fig:vor_b} show
qualitative results for
two instances of these simulations in the form of Voronoi diagrams
for 20 and 40 robots assigned into 3 teams.
While we show the boundaries between individual robots,
our approach assigns teams that cover larger
regions together, indicated by the bold magenta Voronoi lines.
Figures \ref{fig:obs_b} and \ref{fig:obs_c} show team assignment results in the
presence of obstacles, with walls indicated by the gray lines.
These obstacles alter the spatial and communication relationships
among robots, i.e., robots separated by a wall cannot communicate.
It can be seen that our approach
is able to assign teams which are not separated by walls which would
interfere with cooperative operation.

\begin{figure*}[!t]
    \centering
    \subfigure[2 Regions]{
        \label{fig:robots2}
        \centering
        \includegraphics[height=1.0in]{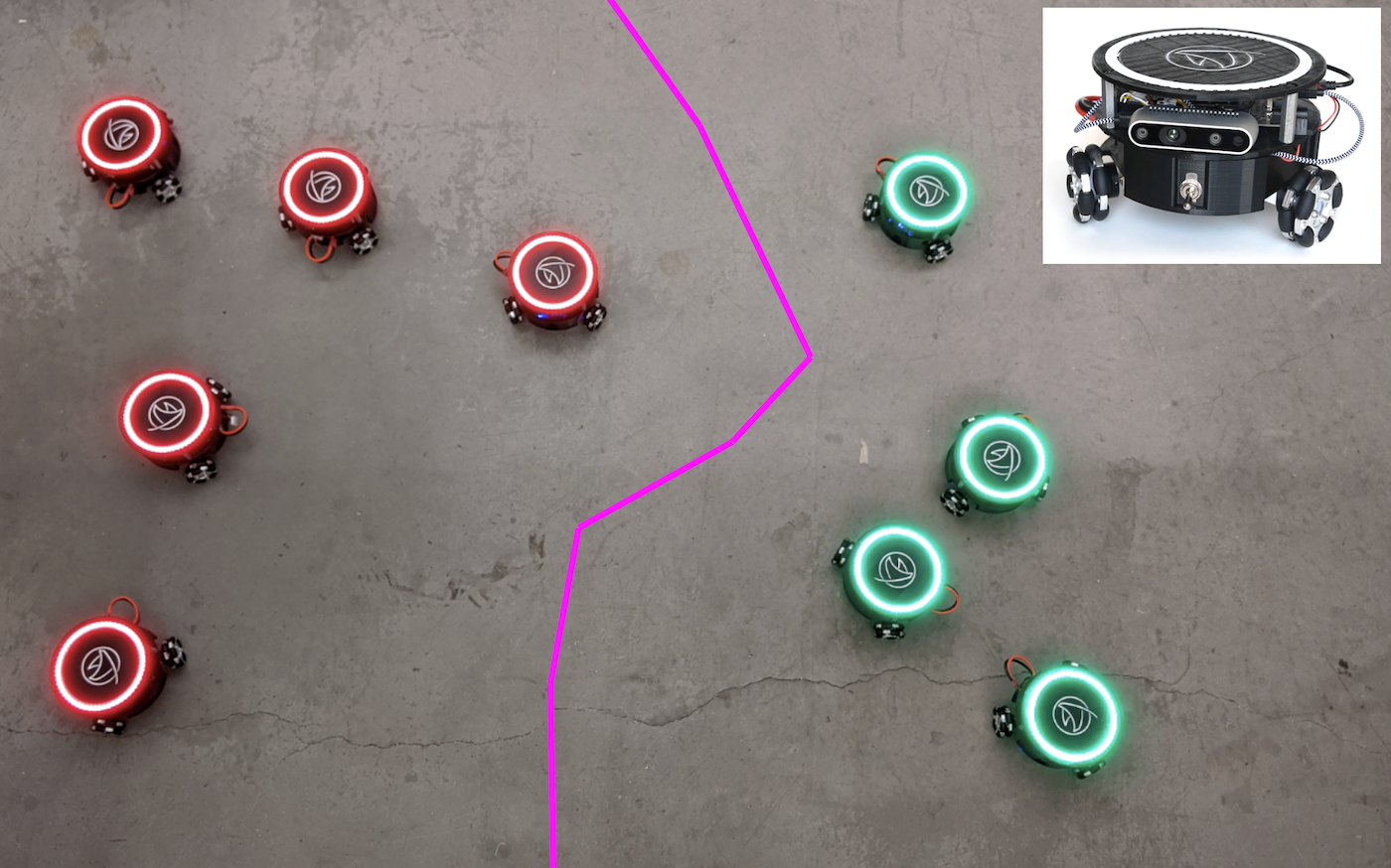}
    }%
    \subfigure[3 Regions]{
        \label{fig:robots3}
        \centering
        \includegraphics[height=1.0in]{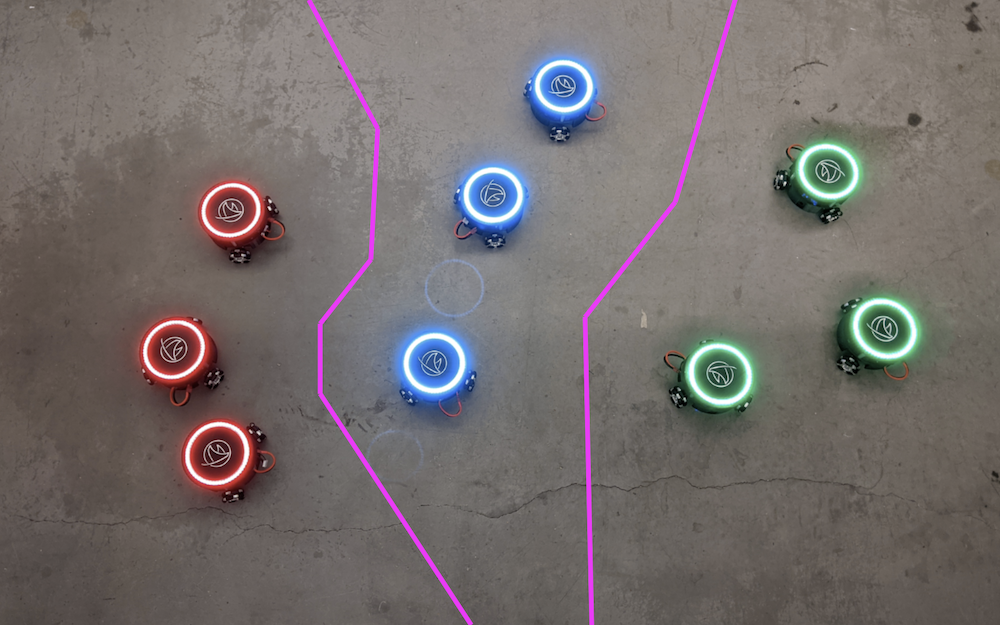}
    }%
    \subfigure[4 Regions]{
        \label{fig:robots4}
        \centering
        \includegraphics[height=1.0in]{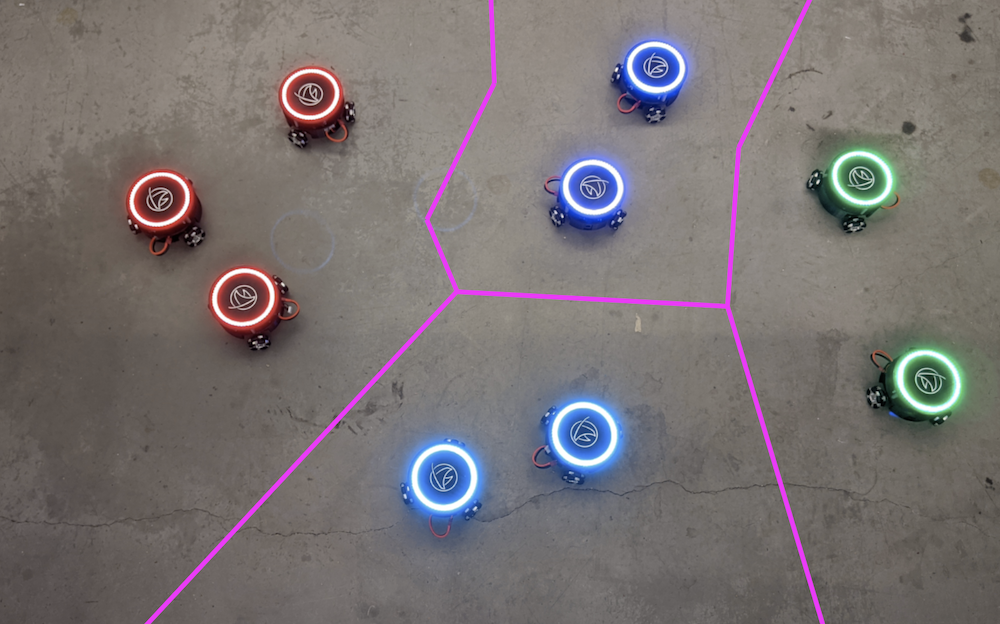}
    }%
    \subfigure[Quantitative]{
        \label{fig:robot_graph}
        \centering
        \includegraphics[height=1.0in]{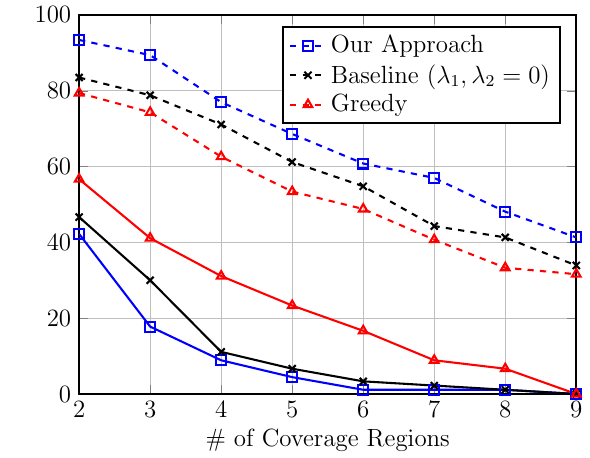}
    }
    \vneg
    \caption{Results from our evaluation on physical
    robots.
    Figures \ref{fig:robots2}--\ref{fig:robots4} show robots
    assigned to teams to cover 2, 3, and 4 regions.
    The magenta lines indicate the separation between
    Voronoi regions that each team
    of robots is assigned to cover, while the LED rings on the
    robots indicate teams.
    Figure \ref{fig:robot_graph} shows the event detection
    (dashed lines) and robot duplication (solid line) rates for
    the physical robot system.}
    \label{fig:robots}
    \vneg
\end{figure*}

Figure \ref{fig:ed_sd} shows the results for the event detection
metric for a subset of the simulations, specifically
$N = \{20, 40\}$ and $| \CCC | = \{3, 5\}$
and covering between 2 and 10 regions.
We see that in most cases, the baseline version of our approach
outperforms the greedy algorithm based only on spatial
relationships.
However, our approach consistently outperforms both,
showing that the block structure induced in the unified
representation matrix corresponds well to teams with high
event detection rates.
Although not displayed due to space limitations, these results
are consistent with those for other values of $N$ 
and $| \mathcal{C} |$.

Figure \ref{fig:ed_sd} also shows results from
the robot duplication metric.
We observe that in all tested cases in the experiments, 
our approach outperforms the greedy approach based purely upon 
distance between robots as well
as a baseline version of our approach.
The teams identified by our approach contain consistently fewer robots
with duplicate sensing capabilities,
demonstrating that our approach is effective at distributing
capabilities among teams.

\subsection{Results on Physical Multi-Robot Systems}

Our second experiment implements our approach on physical robots 
using our multi-robot platform, consisting of $N = 9$ robots.
Each robot, seen in the top corner of Figure \ref{fig:robots2},
is equipped with an RGB camera, depth camera, and microphone
for $| \mathcal{C} | = 3$ sensing capabilities, as well as an 
LED ring for 
status indication.
We randomly disabled sensing capabilities on the robots, so that
each individual robot was limited to a single capability.
Robots moved randomly, with team assignments and simulated events
occurring at discrete time steps.
Figures \ref{fig:robots2}--\ref{fig:robots4}
show sample team assignments for 2 to 4 regions.

Figure \ref{fig:robot_graph} shows the results of this
evaluation in terms of event detection and robot duplication.
We see that our full approach
outperforms the others by a greater margin that on the
larger simulated systems, suggesting that the heterogeneous
relationships our approach learns from are especially influential
on smaller scales.
In terms of robot duplication, our approach again
outperforms the baseline approach and
the greedy approach.
We note that the baseline approach very closely tracks the
performance of our full approach when the number of regions
is four or greater, suggesting that the block structure induced 
by our full approach has a limited influence as the blocks
grow smaller (i.e., the number of robots on each team decreases).
We also see that all three approaches converge 
at duplication of $0\%$ when the number of teams equals
the number of robots; when only
a single robot is present in each region, there can be no duplicates.

\subsection{Hyperparameter Analysis}

\begin{figure}[h]
    \centering
    \subfigure[Event Detection]{
        \label{fig:ternary1_ed}
        \centering
        \includegraphics[width=0.2\textwidth]{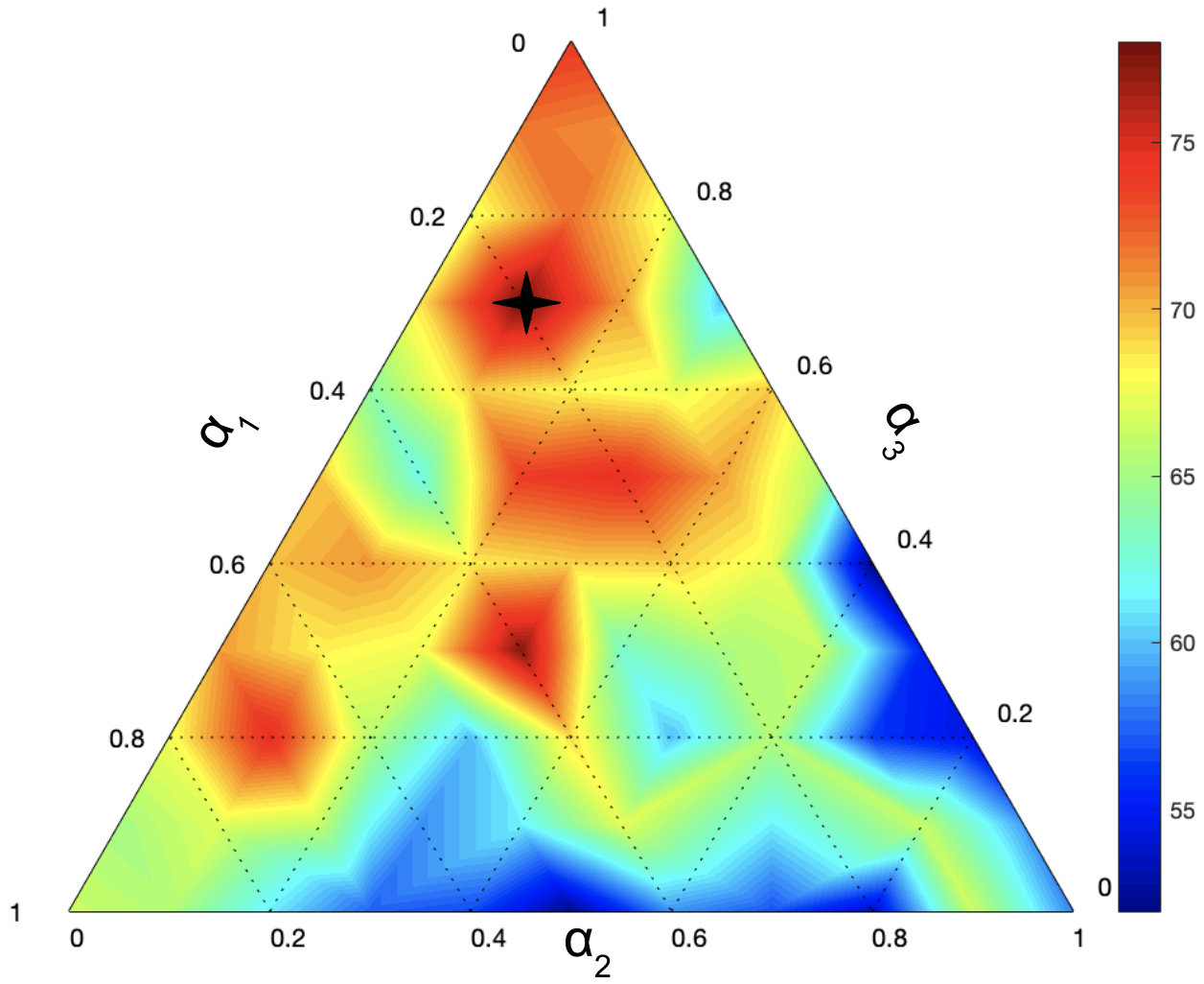}
    }%
    \subfigure[Robot Duplication]{
        \label{fig:ternary1_sd}
        \centering
        \includegraphics[width=0.2\textwidth]{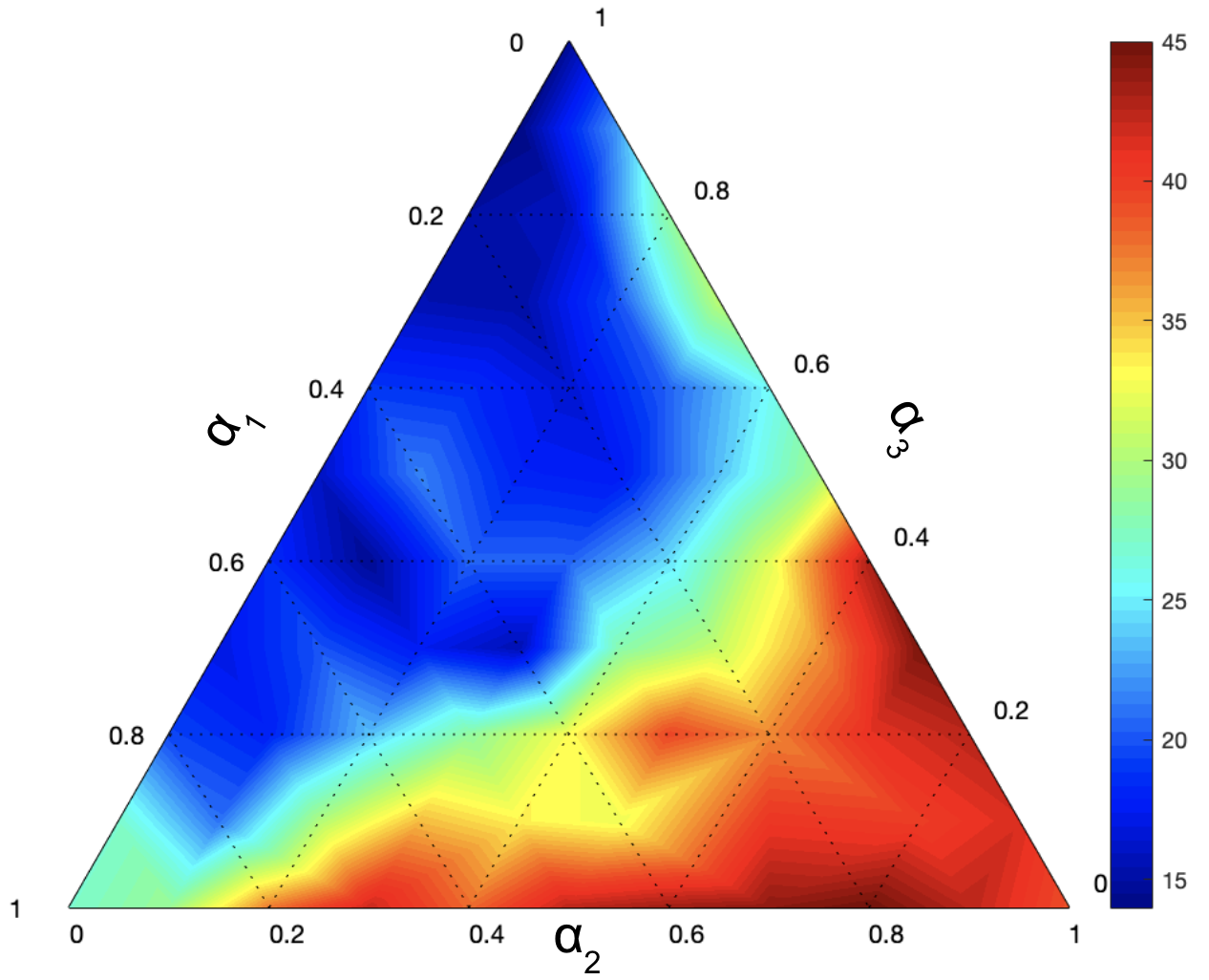}
    }
    \vneg
    \caption{Evaluation of the effect of hyperparameters
    on the event detection
    and robot duplication rates.
For event detection, a high rate (in red) is
good, whereas for robot duplication, a low rate (in blue) is better.
    }
    \label{fig:ternary1}
    \vneg
\end{figure}

We analyze the influence of the hyperparameters
that control the importance of each input graph,
including 
$\alpha_1$ (controlling the importance of the spatial graph), 
$\alpha_2$ (the communication graph), and 
$\alpha_3$ (the sensing capability graph).
Figure \ref{fig:ternary1} shows the 
event detection and robot duplication rates on 
a simulated multi-robot system
resulting from various combinations of the hyperparameters,
in the triangular topological space \cite{zhang2014simplex}
with each side of the triangle corresponding to an $\alpha_i \in [0,1]$
and  $\sum_{i=1}^3 \alpha_i = 1$ with 2 independent values out of 3. 
For example, the black cross in Figure \ref{fig:ternary1_ed}, marking
the best event detection rate,
represents $\alpha_1 = 0.2$, $\alpha_2 = 0.1$, and $\alpha_3 = 0.7$.


In Figure \ref{fig:ternary1_ed}, the event detection is at
its highest for high values of $\alpha_3$, demonstrating
the importance of the sensing capabilities graph.
We observe that balancing $\alpha_1$ and $\alpha_2$ in the
bottom right corner can maintain middling performance, but
relying solely on either of them causes event detection rates
to fall to their lowest values below $55\%$.
Figure \ref{fig:ternary1_sd} shows similar but much 
more consistent effects when evaluating
the rate of robot duplication, with high values of $\alpha_3$ ($ > 0.6$)
corresponding to low rates of duplication ($< 20\%$).
Furthermore, we see that high values of $\alpha_2$ cause higher rates
of duplication, suggesting that grouping nearby robots
is a poor way to ensure heterogeneous teams.

\section{Conclusion}
\label{sec:conclusion}

As real-world robots in a multi-robot system are typically limited in their individual sensing
capabilities, coverage of an area by a heterogeneous multi-robot system
requires the
effective assignment of teams that contain a variety of capabilities.
We propose an approach
to learn a unified representation
of the heterogeneous relationships in the system,
utilizing 
natural divisions to assign teams.
We show that our approach identifies teams with high rates of 
event detection and low duplication of robot capabilities.

\bibliographystyle{ieeetr}
\bibliography{references}

\end{document}